\documentclass{article}

\usepackage{PRIMEarxiv}

\usepackage{amsmath}
\usepackage{amssymb} 
\usepackage{hyperref}
\hypersetup{hidelinks}
\usepackage{natbib}

\usepackage{booktabs}
\usepackage{graphicx}
\usepackage{wrapfig}

\usepackage{multirow}
\usepackage{pifont}
\usepackage{array}
\newcommand{\cmark}{\ding{51}}
\newcommand{\xmark}{\ding{55}}

\usepackage[utf8]{inputenc} 
\usepackage[T1]{fontenc}    
\usepackage{hyperref}       
\usepackage{url}            
\usepackage{booktabs}       
\usepackage{amsfonts}       
\usepackage{nicefrac}       
\usepackage{microtype}      
\usepackage{lipsum}
\usepackage{fancyhdr}       
\usepackage{graphicx}       
\graphicspath{{media/}}     

\title{QuantaSpike: Short-Window Spike-Driven Quantization for Large Language Models
}

\author{
  Bang Hu, Guowei Zhu, Changze Lv, Xiaoqing Zheng, Fengzhe Zhang, Fan Zhang, Wei Cao \\
School of Computer Science, Fudan University, Shanghai, China\\
}

\begin{document}
\maketitle

\begin{abstract}
Large language models (LLMs) achieve strong performance across many tasks but rely on dense multiply-accumulate (MAC) operations during inference, resulting in high energy cost.
Spiking neural networks (SNNs) offer an event-driven alternative in which synaptic integration uses lightweight accumulation.
However, spike-driven LLM inference remains difficult because outlier-heavy activations typically require long firing windows or auxiliary non-spiking paths.
We propose QuantaSpike, a short-window spike-driven quantization framework for LLMs built around Logarithmic Ternary Integrate-and-Fire (LTIF) neurons.
LTIF uses ternary events with power-of-two membrane-response quanta, improving the information represented by each firing step while retaining shift-ACC-compatible computation. 
QuantaSpike combines this neuron with group-adaptive gain and selective outlier admission: normal values use residual LTIF steps, whereas admitted outliers receive one additional onset spike before entering the same residual dynamics.
Across OPT and Llama-2, QuantaSpike achieves state-of-the-art or competitive perplexity and zero-shot accuracy among spike-driven LLM quantization methods. It also transfers to newer dense LLMs, remaining close to the FP16 reference on Llama-3-8B and Qwen3-8B under the same four-step firing window.
Analytical linear-energy projections show that QuantaSpike reduces the energy of one linear transformation by about $80.0\%$ on OPT models and $67.1\%$ on Llama-2 models relative to SpikeQuant, providing an accurate and energy-efficient spike-driven path for LLM inference.
\end{abstract}

\section{Introduction}

\begin{wrapfigure}{R}{0.48\textwidth}
\centering
\includegraphics[width=\linewidth]{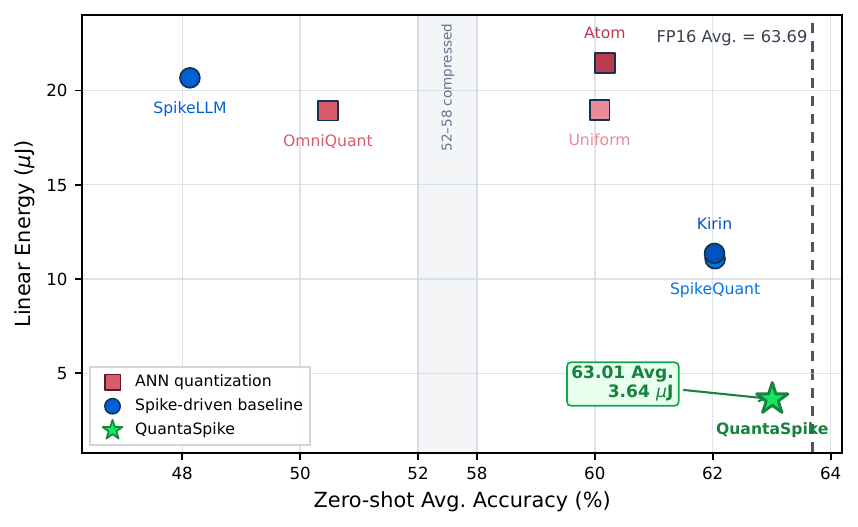}
\caption{
Zero-shot accuracy versus linear energy on Llama-2-7B.
}
\label{fig:overview}
\end{wrapfigure}

In recent years, large language models (LLMs) have become a central foundation for modern Artificial Intelligence systems, driving advancements in natural language understanding, complex reasoning, code generation, and autonomous tool use~\citep{brown2020language,ouyang2022training,chowdhery2022palm,schick2023toolformer,touvron2023llama}.
Although these advances, LLM inference still relies on massive dense multiply-accumulate (MAC) operations, making energy-efficient deployment a persistent challenge.
Spiking neural networks (SNNs), inspired by the brain's event-driven computation, provide an alternative: they carry information through sparse spike events and replace dense MAC operations with lightweight accumulate (ACC) operations, offering high biological plausibility and natural hardware efficiency~\citep{horowitz2014,wang2025spikequant}.
This makes spike-driven LLM inference an appealing direction for reducing dense arithmetic in LLM deployment.
However, representing LLM activations with spikes is difficult: LLM activations often exhibit heavy-tailed distributions and pronounced outliers that are critical to model quality~\citep{dettmers2022llmint8,xiao2023smoothquant}.
Existing spike-driven LLM methods address this tension in two ways: extending the temporal firing window to improve activation fidelity~\citep{xing2025spikellm,wang2025spikequant}, or introducing auxiliary integer-valued paths to handle activations that spikes cannot reliably represent~\citep{wang2026kirin}.
However, the former increases inference latency and spike count, while the latter compromises the purity of spike-driven computation.
The key challenge is therefore to maintain accuracy under a short firing window, while keeping the execution path spike-driven and shift-ACC compatible.

To address this challenge, we propose QuantaSpike, a short-window spike-driven quantization framework for LLMs.
At its core is the Logarithmic LTIF neuron, which emits ternary spike values and uses power-of-two membrane-response quanta to represent values.
QuantaSpike first determines onset admission using both statistical deviation and activation magnitude, then estimates group-adaptive gains from non-admitted values and initializes the membrane states according to the admission decisions.
Normal values are represented by at most three residual LTIF firing steps, while admitted outliers obtain additional dynamic range through a single adaptive onset spike before entering the same residual LTIF dynamics.
Within a firing window of at most four timesteps, the resulting events enable linear-layer computation through sign-controlled shifts and accumulation, without activation-side MAC operations.
Across OPT, Llama-2, Llama-3, and Qwen3 models, QuantaSpike achieves competitive perplexity and zero-shot accuracy, while energy analysis shows that its short firing window and zero-spike silence reduce the number of ACC events actually executed.Our contributions are summarized as follows:
\begin{itemize}
    \item We propose a short-window spike-driven quantization aiming to preserve both normal values and outliers without long firing windows or auxiliary non-spiking computation paths.

    \item We introduce QuantaSpike, a post-training framework centered on the LTIF neuron.
    LTIF combines ternary spike decisions with power-of-two membrane-response quanta within a four-step firing window.

    \item We achieve a strong accuracy--energy trade-off across OPT, Llama-2, Llama-3, and Qwen3 models.
        QuantaSpike delivers state-of-the-art spike-driven accuracy while substantially cutting energy via shift-ACC execution and zero-spike silence under the same analytical energy model used by prior work.
\end{itemize}

\section{Related Work}

\textbf{LLM Quantization.}
Model quantization reduces the precision of weights and activations to lower the
memory and computation costs of LLM inference.
Post-training quantization (PTQ) has become a practical choice for LLMs, including
weight-only methods such as GPTQ~\citep{frantar2023gptq} and
AWQ~\citep{lin2024awq}, as well as weight--activation quantization methods such as
SmoothQuant~\citep{xiao2023smoothquant}, RPTQ~\citep{yuan2023rptq}, and
OmniQuant~\citep{shao2024omniquant}.
Quantization-aware training further improves low-bit accuracy~\citep{liu2024llmqat},
while recent rotation-based methods such as QuaRot~\citep{ashkboos2024quarot} and
SpinQuant~\citep{liu2025spinquant} alleviate activation outliers for more aggressive
quantization.
Despite these advances, quantized LLMs still rely on dense low-bit activation
representations and integer arithmetic.

\textbf{Spike-Driven LLM Inference.}
Spike-driven LLMs aim to reduce the dense arithmetic cost of LLM inference by representing intermediate values with spike events and executing them through event-driven integration.
Existing quantization-based approaches to spike-driven LLM inference broadly fall into two categories: temporal spike encoding and hybrid integer–spike representations.
The former line of work follows temporal spike coding, where LLM activations are encoded into spike sequences and the resulting spiking dynamics are calibrated to approximate pretrained language models~\citep{xing2025spikellm,wang2025spikequant,chen2025las,chen2025fas}.
SpikeLLM uses generalized integrate-and-fire neurons and saliency-based spike allocation to reduce spike length, while SpikeQuant adopts TTFS-based activation encoding with salient-value handling.
LAS introduces specialized neuron representations for outliers and nonlinear operators, while FAS uses coarse-to-fine calibration to reduce conversion error.
These methods improve spike-based LLM inference, but their representation accuracy remains coupled to the firing window or to additional neuron-level encoding complexity.

The latter line of work relaxes the purely spike-driven representation by introducing hybrid precision for difficult activations~\citep{wang2026kirin}.
Kirin observes that compact spike sequences struggle to preserve outlier activations and uses mixed-precision integer-spike representations to improve reconstruction.
This strategy reduces the burden on temporal spike coding, but weakens the uniform spike-driven execution path.
QuantaSpike explores a different point: LTIF maintains accuracy under a short firing window, while keeping the runtime activation representation compatible with shift-ACC execution.

\section{Preliminaries}

Integrate-and-fire (IF) neurons accumulate input current without a leak term and emit a spike when the membrane potential reaches a firing threshold.
Their discrete-time dynamics are written as
\begin{equation}
\left\{
\begin{array}{@{}l@{}}
s^t=\mathbb{I}\!\left(H^{t-1}+I^t\geq\theta\right),\\[1mm]
H^t=H^{t-1}+I^t-s^t\theta.
\end{array}
\right.
\label{eq:if_dynamics}
\end{equation}
Here $I^t$ is the input current, $\theta$ is the firing threshold, and $H^{t-1}$ and $H^t$ are the membrane states before and after the update.
The first equation determines spike emission, while the second describes input integration and the membrane reset following a spike.

Some neurons also form autaptic connections, in which an axon terminal makes a synaptic contact with the same neuron's soma or proximal dendrites.
For example, autaptic feedback is common in cortical GABAergic fast-spiking basket cells and has also been observed in a subset of layer-5 pyramidal cells~\citep{vanderloos1972autapses,bacci2003functional}.
Neurotransmitter release at this self-synaptic contact generates an excitatory or inhibitory postsynaptic response and changes the membrane state of the same neuron after firing.

The spike response model (SRM) abstracts this temporal process by representing the membrane potential as the combination of an input-driven component and a response generated by the neuron's previous spikes. This formulation provides the basis for describing the spike-triggered membrane responses of LTIF~\citep{gerstner1995time,gerstner2014neuronal}.

\section{Method}
\begin{figure*}[t]
      \centering
      \includegraphics[width=\textwidth]{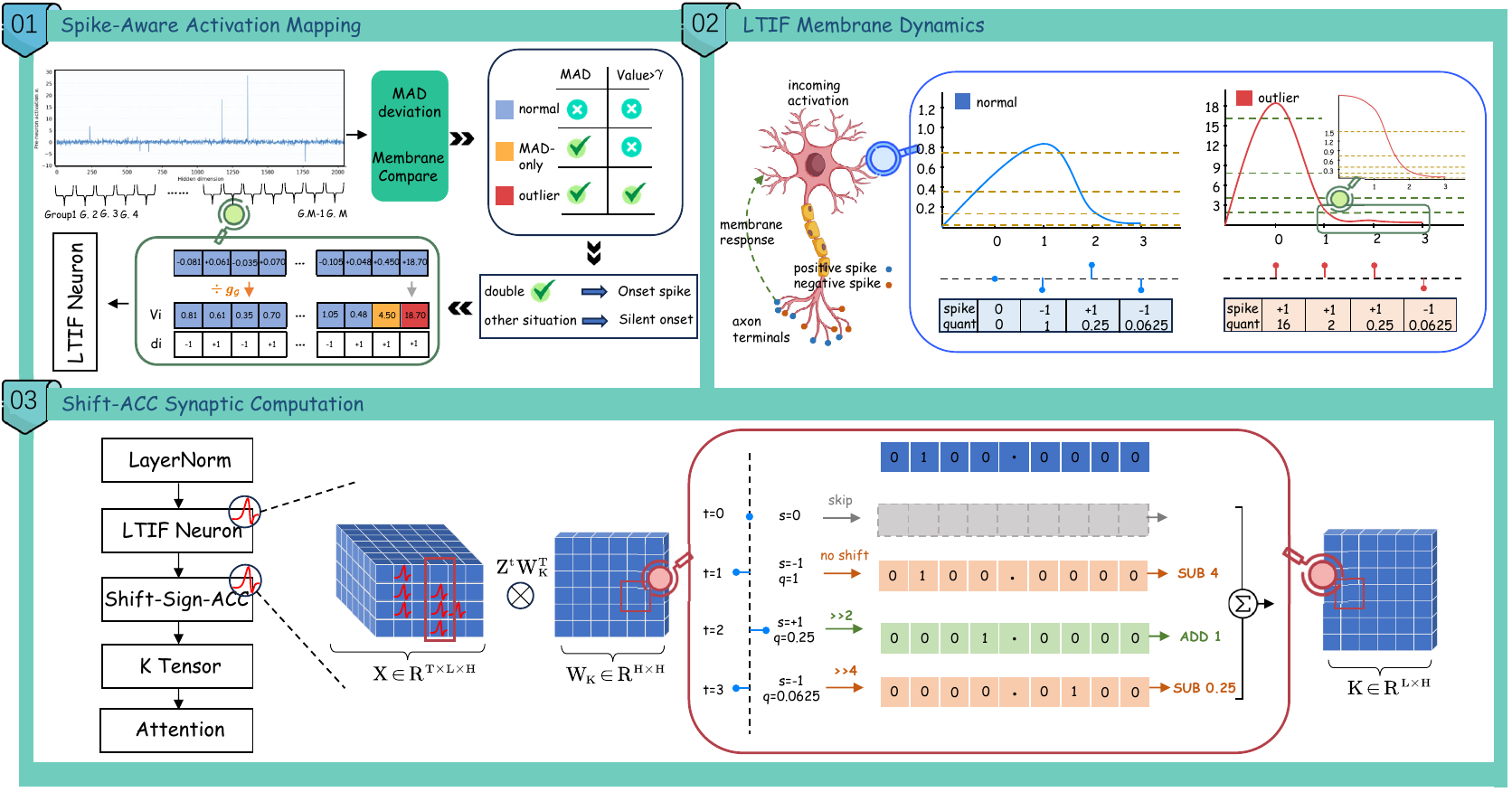}
      \caption{
      \textbf{Overview of QuantaSpike.}
      QuantaSpike maps LLM activations into the LTIF membrane domain, admits only necessary outlier states to an adaptive onset spike, and represents activations with short-window
      ternary LTIF spikes.
      These spikes are processed via sign selection, power-of-two shifts, and accumulation, ensuring a purely spike-driven linear path free of MAC operations.
      }
      \label{fig:method_overview}
\end{figure*}

In this section, we introduce QuantaSpike, a spike-driven activation quantization workflow built around Logarithmic Ternary Integrate-and-Fire (LTIF) neurons.
Figure~\ref{fig:method_overview} gives an overview of the proposed workflow.
To avoid the long firing windows or mixed non-spiking paths used by prior spike-driven LLM quantization methods, QuantaSpike builds the activation representation around LTIF neurons.
We first describe spike-aware activation mapping and the admission rule in Section~\ref{sec:method_admission}, then formulate the LTIF membrane dynamics in Section~\ref{sec:method_ltif}.
Section~\ref{sec:method_shiftacc} shows how the resulting ternary spikes are executed through shift-ACC accumulation.

\subsection{Spike-Aware Activation Mapping}
\label{sec:method_admission}

As shown in Fig.~\ref{fig:method_overview}, QuantaSpike maps LLM activations into initial membrane states for LTIF firing.
The mapping combines two-criterion onset admission with group-wise scaling, allowing ordinary values and admitted outliers to enter the neuron with different dynamic-range requirements.

Given a token activation vector $x\in\mathbb{R}^{H}$, we partition the hidden dimension into contiguous groups of size $G$:
\begin{equation}
    \mathcal{G}_m
    =
    \{j \mid mG \leq j < \min((m+1)G,H)\}.
\end{equation}
Grouping preserves the original channel order and defines the scope of the local quantization scales.

Unlike prior spike-driven methods, we use a strict onset admission rule based jointly on statistical deviation and activation magnitude.
For each token, we compute the median and median absolute deviation over the hidden dimension:
  \begin{equation}
      \mu=\operatorname{median}_{j}(x_j),
      \qquad
      D=\operatorname{median}_{j}|x_j-\mu|.
  \end{equation}
An activation becomes a candidate when its normalized deviation exceeds
  $\lambda$:
  \begin{equation}
      r_i=
      \mathbb{I}\!\left[
      \frac{|x_i-\mu|}
      {1.4826\max(D,\epsilon)}
      >\lambda
      \right].
  \end{equation}
We admit the candidate for onset firing only if its magnitude also exceeds $\gamma$:
  \begin{equation}
      a_i=r_i\,\mathbb{I}[|x_i|>\gamma].
      \label{eq:admission_gate}
  \end{equation}
This second criterion prevents small but statistically unusual values from receiving an unnecessary onset event.

For each token and group $\mathcal{G}_m$, let
$\mathcal{N}_m=\{j\in\mathcal{G}_m:a_j=0\}$ and
$\mathcal{O}_m=\{j\in\mathcal{G}_m:a_j=1\}$, with
$\max\varnothing=0$. We compute path-specific gains as
\begin{equation}
g_m^{p}(x)
=
\max\left(
\frac{\max_{j\in\mathcal{S}_m^{p}}|x_j|}{Q_p},
\epsilon
\right),
\quad
\mathcal{S}_m^{\mathrm n}=\mathcal{N}_m,\quad
\mathcal{S}_m^{\mathrm o}=\mathcal{O}_m,
\quad
Q_p=2^{b_p-1}-1.
\label{eq:group_gain}
\end{equation}
For an admitted outlier, $\widetilde{x}_i=\mathcal{Q}_{\mathrm{o},m}(x_i)$
is reconstructed in the original activation coordinates. The initial membrane
state is normalized according to its path:
\begin{equation}
h_i=
\begin{cases}
x_i/g_{\mathcal{G}(i)}^{\mathrm n}(x), & a_i=0,\\
\widetilde{x}_i/g_{\mathcal{G}(i)}^{\mathrm o}(x), & a_i=1,
\end{cases}
\qquad
u_i^0=h_i .
\label{eq:membrane_init}
\end{equation}

\subsection{Logarithmic Ternary Integrate-and-Fire Neuron}
\label{sec:method_ltif}

Given the membrane initialization and admission decision from Section~\ref{sec:method_admission}, LTIF represents each activation through signed ternary events and power-of-two membrane responses.
The membrane state is initialized as $u_i^0=h_i$, and its dynamics are
  \begin{equation}
  \left\{
  \begin{array}{@{}l@{}}
  s_i^t \in \{-1,0,+1\},\\[1mm]
  u_i^{t+1}
  = u_i^t-s_i^t q_i^t
  = h_i+\displaystyle\sum_{\tau=0}^{t}
  \eta_i(t+1,\tau)s_i^\tau .
  \end{array}
  \right.
  \label{eq:ltif_dynamics}
  \end{equation}
where $s_i^t$ is the ternary firing event and $q_i^t$ is its non-negative response quantum. For the implemented LTIF response,
  \begin{equation}
      \eta_i(t+1,\tau)=-q_i^\tau,
      \qquad 0\leq\tau\leq t.
      \label{eq:ltif_response}
  \end{equation}
Thus, each emitted event introduces a signed membrane correction that is retained in the residual membrane state.
A positive event decreases a positive membrane state, whereas a negative event increases a negative membrane state toward zero. 
The membrane direction can reverse after an update when the residual crosses zero.The admission indicator directly gates the optional onset event: only values with
$a_i=1$ emit at $t=0$, whereas all values enter the shared residual dynamics for
$t\geq1$.

LTIF neurons have three defining properties—signed ternary firing, graded thresholds, and admission-gated onset firing—which together enable them to represent the full activation range within a four-step firing window.

\textbf{Signed ternary firing.}
LTIF emits a signed ternary spike
\begin{equation}
        s_i^t \in \{-1,0,+1\},
        \label{eq:ternary_spike}
\end{equation}
where $+1$ and $-1$ denote positive and negative spike events, respectively, and $0$ denotes a silent timestep.
The sign records the direction of the current membrane state, while the effective spike magnitude is specified by an associated response factor.
Each spike event is therefore paired with a non-negative power-of-two response magnitude:
\begin{equation}
        q_i^t \in \{0\}\cup\{2^{e}:e\in\mathcal{E}\},
        \label{eq:response_quantum}
\end{equation}
where $\mathcal{E}\subset\mathbb{Z}$ is the supported exponent set.
The ternary variable $s_i^t$ determines the sign of the event, whereas $q_i^t$ determines the strength of its membrane-response and its power-of-two contribution during synaptic computation.

\textbf{Graded thresholds.}
LTIF uses a set of graded thresholds to control the response magnitude associated with each ternary spike.
For group $g$ and timestep $t$, calibration constructs an ordered threshold--response map
\begin{equation}
        \mathcal{P}_{g}^{t}
        =
        \left\{
        \bigl(\theta_{g,k}^{t},q_{g,k}^{t}\bigr)
        \right\}_{k=1}^{K_t},
        \qquad
        0<\theta_{g,1}^{t}<\cdots<\theta_{g,K_t}^{t},
        \quad
        q_{g,k}^{t}=2^{e_{g,k}^{t}} .
        \label{eq:threshold_quantum_map}
\end{equation}
The thresholds determine the firing level reached by the membrane, while the corresponding response magnitude determine the strength of the signed membrane-response.

At time step $t$, following the LTIF membrane dynamics, the membrane magnitude $V_i^t=|u_i^t|$ is compared with the ordered thresholds, starting from the largest threshold.
The highest threshold reached by the membrane determines the response level:
\begin{equation}
      k_i^t
      =
      \max
      \left\{
      k \,\middle|\,
      V_i^t\geq\theta_{\mathcal{G}(i),k}^{t}
      \right\}.
      \label{eq:ltif_level_selection}
\end{equation}
where $d_i^t=\operatorname{sgn}(u_i^t)$ (with $d_i^t=0$ when $u_i^t=0$), and the set is empty when $V_i^t<\theta_{\mathcal{G}(i),1}^{t}$.
When $V_i^t$ reaches at least the smallest threshold, LTIF emits a ternary spike
in the current membrane direction and activates the corresponding signed
membrane-response:
\begin{equation}
      (s_i^t,q_i^t)
      =
      \begin{cases}
      \left(d_i^t,q_{\mathcal{G}(i),k_i^t}^{t}\right),
      & V_i^t\geq\theta_{\mathcal{G}(i),1}^{t},\\
      (0,0),
      & V_i^t<\theta_{\mathcal{G}(i),1}^{t}.
      \end{cases}
      \label{eq:ltif_event_selection}
\end{equation}

\textbf{Onset firing.}
According to the admission decision, LTIF actively controls the first timestep by either suppressing the onset event or emitting an adaptive onset spike.
At $t=0$, values with $a_i=0$ are forced to remain silent.
An admitted outlier receives one adaptive onset event:
  \begin{equation}
  s_i^0=a_i\operatorname{sgn}(h_i),
  \qquad
  q_i^0=a_i2^{e_i^0},
  \qquad
  e_i^0=\arg\min_{e\in\mathcal{E}_{\mathrm o}}
  \left|\,|h_i|-2^e\,\right|,
  \label{eq:onset_alignment}
  \end{equation}
Here $\mathcal{E}_{\mathrm o}$ is the exponent set of the coarse outlier
power-of-two bases after normalization by $g^{\mathrm o}_{\mathcal{G}(i)}(x)$;
the bases are constructed from the calibrated outlier range. Thus, $q_i^0=0$ for
non-admitted values. The onset event expands the representable range for
admitted outliers, after which they enter exactly the same graded residual LTIF
dynamics as normal values.
The normalized spike-domain representation and its activation reconstruction are
  \begin{equation}
  z_i=\sum_{t=0}^{n}s_i^tq_i^t,
  \qquad
        \hat{x}_i=g_i(x)z_i,
        \qquad
        g_i(x)=
        \begin{cases}
        g^{\mathrm n}_{\mathcal{G}(i)}(x),&a_i=0,\\
        g^{\mathrm o}_{\mathcal{G}(i)}(x),&a_i=1.
        \end{cases}
      \label{eq:ltif_reconstruction}
  \end{equation}
Normal values are represented using at most $n$ residual firing steps, whereas admitted outliers use one additional onset step. 
With $n=3$, QuantaSpike therefore uses at most four firing steps without introducing a separate non-spiking outlier path.

\subsection{Shift-ACC Computation Analysis}
\label{sec:method_shiftacc}

We now examine how LTIF-coded activations contribute to a linear
transformation. For channel $i$, LTIF produces a normalized representation
\begin{equation}
z_i=\sum_{t=0}^{n}s_i^t q_i^t,
\qquad
s_i^t\in\{-1,0,+1\},
\quad
q_i^t\in\{0\}\cup\{2^{e_i^t}\}.
\label{eq:spike_domain_value}
\end{equation}
The reconstructed activation is $\hat{x}_i=g_i(x)z_i$, where
$g_i(x)=g_{\mathcal{G}(i)}^{\mathrm n}(x)$ for a normal value and
$g_i(x)=g_{\mathcal{G}(i)}^{\mathrm o}(x)$ for an admitted outlier.

For each group $\mathcal{G}_m$, we accumulate the events from the normal
and outlier paths separately:
\begin{equation}
A_m^{p}(x)=
\sum_{\substack{i\in\mathcal{G}_m:\,a_i=p\\t:\,s_i^t\neq0}}
s_i^t\,
\operatorname{Shift}\!\left(W_{:,i},e_i^t\right),
\qquad p\in\{0,1\},
\label{eq:group_event_accumulator}
\end{equation}
where $\operatorname{Shift}(W_{:,i},e_i^t)$ denotes multiplication by
the power-of-two response quantum $2^{e_i^t}$ using the corresponding
binary shift in a suitable fixed-point implementation. A positive event
adds the shifted weight column, a negative event subtracts it, and a
zero event triggers no accumulation. The linear output is reconstructed as
\begin{equation}
\hat{y}
=
\sum_m
\left[
g_m^{\mathrm n}(x)A_m^0(x)
+
g_m^{\mathrm o}(x)A_m^1(x)
\right].
\label{eq:shift_acc_event}
\end{equation}

The gains are computed from the current token and therefore cannot be
folded into fixed weight columns before inference. Instead, the normal
and outlier accumulators are rescaled by their respective group gains
at runtime. Thus, the sign-controlled shift-ACC operations describe
the LTIF event-accumulation path; computing the admission decisions and
gains, followed by group-wise rescaling, incurs additional runtime work.
The operation-level energy projection in Section~\ref{sec:energy_analysis}
isolates the event-accumulation cost and does not include these additional
operations or data movement.

\section{Experiments}

\subsection{Experimental Setup}
\label{sec:exp_setup}

\paragraph{Implementation details.}
QuantaSpike employs a W4A4(4\&5) post-training setting: 4-bit uniform weights and 4-bit activations with a 5-bit outlier budget. By default, we use a four-step firing window ($n=3$ residual LTIF steps and one optional adaptive onset step) with a group size $G=64$. We follow the standard 128-sequence, 2048-token WikiText-2 calibration protocol; the sampling and caching procedure is detailed in Appendix~\ref{app:implementation}. Experiments run on a single NVIDIA A100 80GB GPU.

\paragraph{Evaluation tasks.}
We evaluate QuantaSpike on OPT~\citep{zhang2022opt} (1.3B, 2.7B), Llama-2~\citep{touvron2023llama2} (7B, 13B), Llama-3~\citep{grattafiori2024llama3} (8B), and Qwen3~\citep{yang2025qwen3} (8B). Metrics include language modeling perplexity on WikiText-2~\citep{merity2017pointer} and C4~\citep{raffel2020exploring}, and zero-shot accuracy on PIQA~\citep{bisk2020piqa}, ARC-Easy, ARC-Challenge~\citep{clark2018arc}, HellaSwag~\citep{zellers2019hellaswag}, and WinoGrande~\citep{sakaguchi2020winogrande} following standard protocols.

\paragraph{Baselines.}
For OPT and Llama-2, we compare against ANN PTQ baselines (Vanilla, GPTQ~\citep{frantar2023gptq}, OmniQuant~\citep{shao2024omniquant}, Atom~\citep{zhao2024atom}) and spike-driven methods (SpikeLLM~\citep{xing2025spikellm}, SpikeQuant~\citep{wang2025spikequant}, Kirin~\citep{wang2026kirin}). For Llama-3 and Qwen3, where spike-driven results are limited, we evaluate FP16, uniform W4A4(4\&5), and ANN baselines (GPTQ, AWQ~\citep{lin2024awq}) alongside QuantaSpike under a unified pipeline with identical checkpoints, tokenizers, and evaluation setups.

\subsection{Main Results}
\label{sec:main_results}

In this section, we focus on whether QuantaSpike can maintain accuracy under a short firing window.
Following prior spike-driven LLM studies, we report WikiText-2 and C4 perplexity for token-level language modeling, followed by zero-shot accuracy on five commonsense tasks. 
As shown in Table~\ref{tab:ppl_main}, QuantaSpike uses the shortest firing window among the compared spike-driven methods while maintaining competitive perplexity.
On OPT models, it matches or improves upon SpikeQuant on both datasets and achieves the best WikiText-2 perplexity among quantized baselines on OPT-2.7B.
On Llama-2 models, its perplexity remains close to SpikeQuant and Kirin despite reducing the firing window from 16 or 32 steps to four, indicating that the LTIF onset-and-residual representation preserves language-modeling quality.

\begin{table*}[h]
\centering
\caption{\textbf{Perplexity on WikiText-2(WT-2) and C4 across model scales.}We compare SpikeQuant (SQ), SpikeLLM (SL) and Kirin with our method QuantaSpike(QS)}
\label{tab:ppl_main}
\scriptsize
\setlength{\tabcolsep}{1.8pt}
\renewcommand{\arraystretch}{1.12}
\begin{tabular*}{\textwidth}{@{\extracolsep{\fill}}l*{4}{c}|*{4}{c}|*{5}{c}|*{5}{c}@{}}
\toprule
Metric
& \multicolumn{4}{c|}{OPT-1.3B}
& \multicolumn{4}{c|}{OPT-2.7B}
& \multicolumn{5}{c|}{Llama-2-7B}
& \multicolumn{5}{c}{Llama-2-13B} \\
\cmidrule(lr){2-5} \cmidrule(lr){6-9} \cmidrule(lr){10-14} \cmidrule(lr){15-19}
& FP16 & \shortstack{SQ} & Kirin & \shortstack{QS}
& FP16 & \shortstack{SQ} & Kirin & \shortstack{QS}
& FP16 & \shortstack{SL} & \shortstack{SQ} & Kirin & \shortstack{QS}
& FP16 & \shortstack{SL} & \shortstack{SQ} & Kirin & \shortstack{QS} \\
\midrule
Bits
& \shortstack{W16\\A16} & \shortstack{W4A\\(4\&5)} & \shortstack{W4A\\(4\&8)} & \shortstack{W4A\\(4\&5)}
& \shortstack{W16\\A16} & \shortstack{W4A\\(4\&5)} & \shortstack{W4A\\(4\&8)} & \shortstack{W4A\\(4\&5)}
& \shortstack{W16\\A16} & \shortstack{W4A\\(4\&8)} & \shortstack{W4A\\(4\&5)} & \shortstack{W4A\\(4\&8)} & \shortstack{W4A\\(4\&5)}
& \shortstack{W16\\A16} & \shortstack{W4A\\(4\&8)} & \shortstack{W4A\\(4\&5)} & \shortstack{W4A\\(4\&8)} & \shortstack{W4A\\(4\&5)} \\
Timestep
& -- & 32 & 16 & 4
& -- & 32 & 16 & 4
& -- & 256 & 32 & 16 & 4
& -- & 256 & 32 & 16 & 4 \\
WT-2 $\downarrow$
& 14.63 & \underline{16.38} & \textbf{16.33} & \underline{16.38}
& 12.47 & 13.15 & \underline{12.89} & \textbf{12.76}
& 5.68 & 11.36 & 5.80 & \textbf{5.71} & \underline{5.78}
& 4.88 & 9.71 & 5.04 & \textbf{5.03} & 5.11 \\
C4 $\downarrow$
& 14.72 & 16.56 & \textbf{16.26} & \underline{16.34}
& 13.17 & 13.91 & \textbf{13.72} & \underline{13.79}
& 7.08 & 15.87 & 7.33 & \underline{7.25} & \textbf{7.23}
& 6.46 & 12.10 & 6.65 & \textbf{6.61} & \textbf{6.61} \\
\bottomrule
\end{tabular*}
\end{table*}

Table~\ref{tab:zeroshot_main} further reports zero-shot accuracy on five commonsense tasks.
QuantaSpike achieves the best average score on three of the four model settings, and is only $0.05$ points behind Kirin on OPT-2.7B.
Despite using a four-step firing window, QuantaSpike remains close to the FP16 reference, with only a $0.68$-point average drop on both OPT-2.7B and Llama-2-7B.
It also consistently outperforms the ANN quantization baselines in average accuracy, while preserving a spike-driven execution path.
On Llama-2 models, QuantaSpike is particularly stable across PIQA, ARC-Easy, HellaSwag, and WinoGrande, showing that the shorter firing window does not translate into a systematic loss on downstream tasks.
Together with the perplexity results, these findings support the central claim that QuantaSpike maintains accuracy under a short firing window while keeping activation computation spike-driven.

\begin{table*}[!t]
\centering
\caption{\textbf{Zero-shot accuracy on OPT and Llama-2 models.}
FP16 and QuantaSpike are evaluated in our pipeline; other baselines are reported from prior papers.
Best and second-best results among quantized or spike-driven methods are bolded and underlined.}
\label{tab:zeroshot_main}
\footnotesize
\setlength{\tabcolsep}{2.8pt}
\renewcommand{\arraystretch}{1.04}
\begin{tabular*}{0.90\textwidth}{@{\extracolsep{\fill}}llc*{6}{c}@{}}
\toprule
Model & Method & \shortstack{SNN\\Conv.} & PIQA & \shortstack{ARC-\\Easy} & \shortstack{ARC-\\Challenge} & HellaSwag & WinoGrande & Avg. \\
\midrule
\multirow{6}{*}{OPT-1.3B}
& FP16 & \xmark & 72.52 & 50.84 & 29.78 & 53.72 & 59.75 & 53.32 \\
\cmidrule(lr){2-9}
& Vanilla Quant. & \xmark & 67.79 & 42.93 & 25.26 & 45.94 & 54.93 & 47.37 \\
& GPTQ & \xmark & 68.34 & 46.17 & \underline{27.65} & 46.53 & 55.26 & 48.79 \\
\cmidrule(lr){2-9}
& SpikeQuant & \cmark & 71.49 & \textbf{49.66} & 27.22 & 51.46 & \textbf{58.56} & 51.68 \\
& Kirin & \cmark & \textbf{71.86} & 48.78 & \underline{28.67} & \underline{52.25} & \underline{57.77} & \underline{51.87} \\
& QuantaSpike & \cmark & \underline{71.84} & \underline{49.23} & \textbf{28.79} & \textbf{52.34} & 57.70 & \textbf{51.98} \\
\midrule
\multirow{6}{*}{OPT-2.7B}
& FP16 & \xmark & 74.81 & 54.34 & 31.31 & 60.62 & 61.01 & 56.42 \\
\cmidrule(lr){2-9}
& Vanilla Quant. & \xmark & 71.16 & 50.84 & 28.75 & 55.57 & \textbf{60.46} & 53.36 \\
& GPTQ & \xmark & 71.38 & 48.19 & 27.82 & 49.86 & 54.82 & 50.41 \\
\cmidrule(lr){2-9}
& SpikeQuant & \cmark & 74.21 & 53.79 & \textbf{31.66} & \textbf{59.11} & 59.75 & 55.70 \\
& Kirin & \cmark & \textbf{74.81} & \textbf{53.91} & \underline{31.48} & 58.83 & 59.91 & \textbf{55.79} \\
& QuantaSpike & \cmark & \underline{74.55} & \underline{53.84} & 30.97 & \underline{59.05} & \underline{60.30} & \underline{55.74} \\
\midrule
\multirow{8}{*}{Llama-2-7B}
& FP16 & \xmark & 78.51 & 55.54 & 42.33 & 73.99 & 68.06 & 63.69 \\
\cmidrule(lr){2-9}
& Vanilla Quant. & \xmark & 75.46 & 53.96 & 40.19 & 69.21 & 61.56 & 60.08 \\
& OmniQuant & \xmark & 66.15 & 45.20 & 31.14 & 56.44 & 53.43 & 50.47 \\
& Atom & \xmark & 76.28 & 52.10 & 38.99 & 69.81 & 63.69 & 60.17 \\
\cmidrule(lr){2-9}
& SpikeLLM & \cmark & 64.47 & 48.74 & 27.30 & 43.29 & 56.83 & 48.13 \\
& SpikeQuant & \cmark & \underline{77.09} & \underline{55.39} & 40.36 & 72.09 & 65.27 & \underline{62.04} \\
& Kirin & \cmark & 76.77 & 54.55 & \underline{40.53} & \underline{72.10} & \underline{66.22} & 62.03 \\
& QuantaSpike & \cmark & \textbf{77.91} & \textbf{55.44} & \textbf{41.54} & \textbf{72.52} & \textbf{67.64} & \textbf{63.01} \\
\midrule
\multirow{8}{*}{Llama-2-13B}
& FP16 & \xmark & 80.52 & 57.53 & 44.98 & 77.38 & 70.22 & 66.13 \\
\cmidrule(lr){2-9}
& Vanilla Quant. & \xmark & 76.71 & 55.13 & 41.47 & 73.09 & 65.19 & 62.32 \\
& OmniQuant & \xmark & 69.69 & 47.39 & 33.10 & 58.96 & 55.80 & 52.99 \\
& Atom & \xmark & 77.69 & \textbf{57.58} & 42.92 & 73.77 & 68.51 & 64.09 \\
\cmidrule(lr){2-9}
& SpikeLLM & \cmark & 66.49 & 55.30 & 30.12 & 47.43 & 51.54 & 50.18 \\
& SpikeQuant & \cmark & 79.33 & 56.48 & \textbf{44.03} & 75.53 & 68.11 & 64.70 \\
& Kirin & \cmark & \underline{79.35} & 56.40 & \underline{43.94} & \underline{75.94} & \underline{68.98} & \underline{64.92} \\
& QuantaSpike & \cmark & \textbf{79.54} & \underline{56.63} & 43.18 & \textbf{76.26} & \textbf{69.43} & \textbf{65.01} \\
\bottomrule
\end{tabular*}
\end{table*}

We further evaluate newer dense LLM families in Table~\ref{tab:modern_dense}.
On Llama-3-8B, QuantaSpike obtains $6.98$ WikiText-2 perplexity and a $71.25$ zero-shot average, improving over GPTQ by $0.28$ perplexity points and $1.11$ zero-shot points.
Its zero-shot score is also within $0.09$ points of Uniform and $0.08$ points of AWQ.
On Qwen3-8B, QuantaSpike obtains $10.04$ perplexity and a $69.41$ zero-shot average; its perplexity is within $0.33$ of FP16 and lower than those of AWQ and GPTQ.
Compared with FP16, the zero-shot gaps are $1.48$ points on Llama-3-8B and $2.17$ points on Qwen3-8B. 
These results show that the LTIF representation transfers to newer dense LLMs under the same four-step firing window.

  \begin{table}[!t]
      \centering
      \caption{\textbf{WikiText-2 perplexity and zero-shot accuracy on newer dense LLMs.} 
      Comparison of FP16, Uniform, AWQ, GPTQ and our QuantaSpike.}
      \label{tab:modern_dense}
      \scriptsize
      \setlength{\tabcolsep}{3pt}
      \renewcommand{\arraystretch}{1.12}
      \begin{tabular}{l*{10}{c}}
          \toprule
          \multirow{2}{*}{Method}
          & \multicolumn{5}{c}{Llama-3-8B}
          & \multicolumn{5}{c}{Qwen3-8B} \\
          \cmidrule(lr){2-6}
          \cmidrule(lr){7-11}
          & FP16 & Uniform & AWQ & GPTQ & QuantaSpike
          & FP16 & Uniform & AWQ & GPTQ & QuantaSpike \\
          \midrule
          SNN Conv.
          & \xmark & \xmark & \xmark & \xmark & \cmark
          & \xmark & \xmark & \xmark & \xmark & \cmark \\
          WikiText-2 $\downarrow$
          & 6.14 & 6.88 & 6.55 & 7.26 & 6.98
          & 9.71 & 10.08 & 10.08 & 10.23 & 10.04 \\
          ZS Avg. $\uparrow$
          & 72.73 & 71.34 & 71.33 & 70.14 & 71.25
          & 71.58 & 70.01 & 70.66 & 70.59 & 69.41 \\
          \bottomrule
      \end{tabular}
  \end{table}

\subsection{Shift-ACC Energy Analysis}
\label{sec:energy_analysis}

Each nonzero LTIF event is executed by sign selection, a power-of-two shift, and an ACC update, while silent residual steps trigger no accumulation.
Let $p_o$ denote the fraction of activations that emit an onset spike and $p_z$ the fraction of silent residual steps. 
The average number of nonzero events per activation is
  \begin{equation}
      \bar{m}=n(1-p_z)+p_o .
      \label{eq:event_budget}
  \end{equation}
As detailed in Appendix~\ref{app:firing_budget}, accounting for silence reduces $\bar{m}$ from $3.03$ to $1.34$ on OPT-1.3B and from $3.01$ to $2.08$ on Llama-2-7B.

We estimate the energy of one linear transformation using the operation costs adopted by SpikeQuant and Kirin~\citep{wang2025spikequant,wang2026kirin}.
After normalizing all costs by a 4-bit ACC, the LTIF event cost and the resulting linear energy are
  \begin{equation}
  \left\{
  \begin{array}{@{}l@{}}
  C_{\mathrm{QS}}
  = n(1-p_z)(1+\alpha_{\mathrm{shift}})
    + p_o(\rho_5+\alpha_{\mathrm{shift}}),\\[1mm]
  E_{\mathrm{QS}}^{\mathrm{linear}}(M)
  = E_{\mathrm{SQ}}^{\mathrm{linear}}(M)
    \dfrac{C_{\mathrm{QS}}}{C_{\mathrm{SQ}}(M)}.
  \end{array}
  \right.
  \label{eq:linear_energy_qs}
  \end{equation}
where $\rho_5=E_{\mathrm{ACC}}^{5}/E_{\mathrm{ACC}}^{4}$, $\alpha_{\mathrm{shift}} =E_{\mathrm{shift}}/E_{\mathrm{ACC}}^{4}=0.2$, and $C_{\mathrm{SQ}}(M)$ is the corresponding normalized SpikeQuant cost.

\begin{figure}[t]
      \centering
      \includegraphics[width=\linewidth]{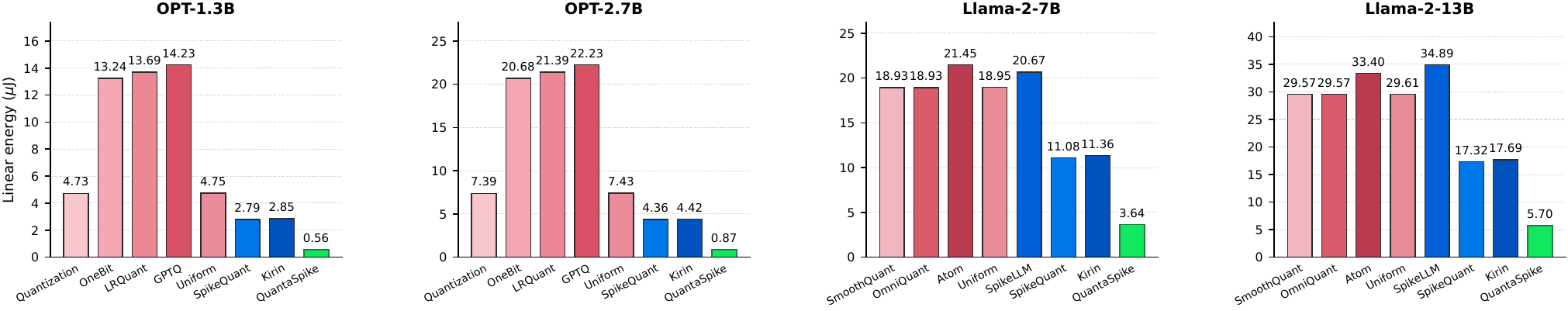}
      \caption{\textbf{Estimated energy of one linear transformation}
      under the operation-cost model adopted by SpikeQuant and Kirin.}
      \label{fig:linear_energy}
  \end{figure}

Figure~\ref{fig:linear_energy} shows that QuantaSpike has the lowest estimated linear-transformation energy across the evaluated OPT and Llama-2 models.
Relative to SpikeQuant, it reduces linear energy by approximately $80.0\%$ on OPT and $67.1\%$ on Llama-2. 
These savings result from replacing dense activation-side MACs with sparse sign-controlled shift-ACC events and skipping accumulation during silent steps.
Appendix~\ref{app:energy} details the operation-cost assumptions and linear-energy projections.

\subsection{Ablation Studies}
\label{sec:ablation}

We ablate two design choices in QuantaSpike: selective onset admission and the onset spike itself.
Removing the magnitude gate leaves MAD-based admission alone, whereas removing the onset spike shortens the firing window from four to three timesteps and forces all values, including outliers, into the residual LTIF dynamics.

\begin{table*}[h]
\centering
\caption{\textbf{Ablation studies.} The onset spike improves outlier reconstruction, while the magnitude gate controls false outlier admission.}
\label{tab:ablation_all}
\footnotesize
\setlength{\tabcolsep}{4.0pt}
\renewcommand{\arraystretch}{1.04}
\begin{tabular*}{\textwidth}{@{\extracolsep{\fill}}lllcccc@{}}
\toprule
Ablation & Model & Variant & PPL & $p_o$ & $\bar{m}$ & Ratio \\
\midrule
\multirow{4}{*}{Onset spike}
& \multirow{2}{*}{OPT-1.3B}
& QuantaSpike & 16.38 & 2.90\% & 1.34 & 0.262 \\
& & w/o onset spike & 18.78 & 0.00\% & 1.18 & 0.230 \\
\cmidrule(lr){2-7}
& \multirow{2}{*}{Llama-2-7B}
& QuantaSpike & 5.78 & 0.77\% & 2.08 & 0.407 \\
& & w/o onset spike & 6.04 & 0.00\% & 1.95 & 0.382 \\
\midrule
\multirow{2}{*}{Magnitude gate}
& \multirow{2}{*}{Llama-2-7B}
& QuantaSpike & 5.78 & 0.77\% & 2.08 & 0.407 \\
& & w/o magnitude gate & 39.47 & 4.40\% & 2.27 & 0.445 \\
\bottomrule
\end{tabular*}
\end{table*}

\paragraph{Selective onset admission.}
As shown in Table~\ref{tab:ablation_all}, MAD-only admission increases the onset-spike rate on Llama-2-7B from $0.77\%$ to $4.40\%$, while degrading WikiText-2 perplexity from $5.78$ to $39.47$.
Statistical deviation alone therefore admits small-magnitude values that consume additional events without improving accuracy.
The magnitude gate prevents these values from entering the onset-enabled representation.

\paragraph{Onset firing.}
Removing the onset spike saves one timestep but eliminates the dedicated response for admitted outliers.
Consequently, WikiText-2 perplexity increases from $16.38$ to $18.78$ on OPT-1.3B and from $5.78$ to $6.04$ on Llama-2-7B, with a larger degradation on OPT-1.3B.
Together, these results support selectively admitting outliers to an onset spike rather than broadly admitting MAD-selected values or eliminating onset firing.

\section{Conclusion}
In this paper, we presented QuantaSpike, a short-window spike-driven quantization framework for LLMs.
QuantaSpike introduces the Logarithmic Ternary Integrate-and-Fire (LTIF) neuron, which combines signed ternary firing with power-of-two membrane-response quanta to increase the representation capacity of each firing step.
Together with group-adaptive gain and selective onset admission, LTIF represents normal values and outliers through shared residual dynamics within a four-step window, while retaining an activation-side shift-ACC path without dense MAC operations.
Across OPT, Llama-2, Llama-3, and Qwen3, QuantaSpike achieves state-of-the-art or competitive perplexity and zero-shot accuracy among spike-driven LLM quantization methods.
Analytical energy estimates further show that QuantaSpike reduces linear-transformation energy by about $80.0\%$ on OPT and $67.1\%$ on Llama-2 relative to SpikeQuant, supporting accurate and energy-efficient spike-driven LLM inference.
The limitations and future directions are discussed in Appendix~\ref{app:limitations_future}.

\section{Reproducibility Statement}
\label{sec:reproducibility}

Appendix~\ref{app:implementation} specifies the quantization settings, calibration data and sampling seed, evaluation protocol, and model-specific hyperparameters. The remaining appendices provide full task results, ablations, firing-event accounting, and the assumptions behind the analytical energy estimates.
QuantaSpike is a post-training method and does not update the pretrained model weights. 
Our reported perplexity and zero-shot scores are single evaluations of each quantized checkpoint under a fixed calibration seed.
FP16, uniform quantization, and QuantaSpike results generated in our pipeline are distinguished from baseline results taken from published reports, which retain the source papers' evaluation settings. We will release the source code, configurations, calibration and evaluation scripts, and plotting scripts upon publication.

\section{Ethics Statement}
\label{sec:ethics}

Our experiments use publicly available pretrained language models and benchmark datasets; this work collects no new human data. We use these resources for research under the terms and documentation provided by their original publishers.
Quantization can make LLM inference less costly, but it does not remove the potential for biased, inaccurate, or harmful model outputs. Applications using quantized models should therefore retain appropriate evaluation and safeguards for their intended setting.
The efficiency results in this paper are operation-level estimates, consistent with the practices of prior works, and should be interpreted within the stated accounting scope.

\bibliographystyle{unsrt}  
\bibliography{references}  

\appendix

\section{Additional Experimental Details}
\label{app:implementation}

\subsection{Quantization and Calibration}

QuantaSpike is applied after pretraining and does not update model weights.
All quantized models use the W4A4(4\&5) setting described in the main text: weights are quantized to 4-bit uniform values, normal activation states use a 4-bit LTIF representation, and admitted outliers use one additional onset firing step, corresponding to a 5-bit representation budget.
The residual LTIF path uses $n=3$ firing steps unless otherwise specified, and the optional onset response increases the maximum firing window to four steps.

\begin{wraptable}{r}{0.55\textwidth} 
\centering
\caption{
\textbf{Model-specific QuantaSpike configurations.}
$n$ is the number of residual LTIF firing steps, $G$ is the group size, $\lambda$ is the MAD coefficient, and $\gamma$ is the magnitude-gate coefficient.
}
\label{tab:appendix_configs}
\scriptsize
\setlength{\tabcolsep}{3.2pt}
\renewcommand{\arraystretch}{1.05}
\begin{tabular}{lccccc}
\toprule
Model family & $n$ & $G$ & $\lambda$ & $\gamma$ & Max. window \\
\midrule
OPT & 3 & 64 & 3.0 & 0.5 & 4 \\
Llama-2 & 3 & 64 & 3.0 & 0.5 & 4 \\
Llama-3 & 3 & 32 & 3.0 & 0.5 & 4 \\
Qwen3 dense & 3 & 32 & 3.0 & 0.6 & 4 \\
Qwen3 MoE & 3 & 32 & 3.0 & 0.6 & 4 \\
\bottomrule
\end{tabular}
\end{wraptable}

For an activation vector, QuantaSpike partitions the hidden dimension into contiguous groups of size $G$ and computes token-dependent normal and outlier gains for each group.
To match the common protocol used by prior data-driven PTQ and spike-driven LLM methods, calibration uses 128 sequences of 2048 tokens randomly sampled from the WikiText-2 training split with seed 0.
The calibration sequences are disjoint from the evaluation split and are tokenized with the corresponding model tokenizer.
Hooks capture the inputs to every target linear layer over these sequences, and the captured tensors are concatenated for calibration.
The calibration activations determine the normal LTIF prototypes and exponent range.
At evaluation time, the median and median absolute deviation (MAD) are recomputed
for each token to identify outlier candidates, and the magnitude gate determines
whether a candidate receives the optional onset spike.

During calibration, we construct the normal and onset exponent sets,
threshold-response maps, reset prototypes, and calibrated onset bases
from the captured layer inputs. These calibration-time quantities are
cached per layer and remain fixed during evaluation. In contrast,
token-dependent medians, MAD statistics, admission masks, and
path-specific gains are recomputed for each input token. QuantaSpike
therefore requires neither task-specific fine-tuning nor gradient
updates. We cache the calibration-time quantities using the model,
layer, number of residual firing steps, group size, bit width, and
exponent range as the cache key. Changing any of these quantities
invalidates the corresponding cache entry.

Table~\ref{tab:appendix_configs} lists the configurations used for the reported results.
OPT and Llama-2 use the default setting.
For Llama-3-8B and Qwen3-8B, a smaller group size better matches their local activation statistics while preserving the same four-step firing window.

\subsection{Evaluation Protocol}

Perplexity is evaluated on the WikiText-2 test split and the English C4 validation split, using each model's tokenizer and non-overlapping 2048-token windows. For C4, we follow the data-construction convention in prior spike-driven LLM evaluation: tokenize the concatenation of the first 1,100 documents from the first English validation shard (\texttt{en/c4-validation.00000-of-00008.json.gz}), retain the first $256\times2048$ tokens, and compute perplexity over the resulting 256 windows. This C4 evaluation data is separate from the WikiText-2 training sequences used for calibration. FP16, uniform quantization, and QuantaSpike are evaluated on the same windows within our pipeline; values copied from prior papers remain under their published protocols.
Zero-shot evaluation is performed with the same version of the LM Evaluation Harness across all methods evaluated in our pipeline.
We report normalized accuracy for PIQA, ARC-Easy, ARC-Challenge, and HellaSwag, and accuracy for WinoGrande.
The reported average is the unweighted mean over these five tasks.
All post-training quantization and evaluation experiments are conducted on a single NVIDIA A100 80GB GPU.

\begin{wraptable}{r}{0.45\textwidth} 
\centering
\caption{
\textbf{Hyperparameter sensitivity on newer dense LLMs.}
The selected four-step configurations are marked in bold.
}
\label{tab:modern_sensitivity}
\scriptsize
\setlength{\tabcolsep}{3.2pt}
\renewcommand{\arraystretch}{1.04}
\begin{tabular}{lcccc}
\toprule
Model & $n$ & $G$ & $\gamma$ & WikiText-2 PPL $\downarrow$ \\
\midrule
\multirow{4}{*}{Llama-3-8B}
& 3 & 64 & 0.5 & 7.211 \\
& \textbf{3} & \textbf{32} & \textbf{0.5} & \textbf{6.979} \\
& 3 & 32 & 0.6 & 7.002 \\
& 4 & 32 & 0.5 & 6.958 \\
\midrule
\multirow{5}{*}{Qwen3-8B}
& 3 & 64 & 0.5 & 10.109 \\
& 3 & 32 & 0.5 & 10.046 \\
& \textbf{3} & \textbf{32} & \textbf{0.6} & \textbf{10.037} \\
& 3 & 32 & 0.7 & 10.055 \\
& 4 & 32 & 0.5 & 10.046 \\
\bottomrule
\end{tabular}
\end{wraptable}

\section{Full Results on Newer Dense LLMs}
\label{app:modern_dense}

Prior spike-driven LLM studies primarily report results on OPT and Llama-2.
We therefore evaluate Llama-3-8B and Qwen3-8B against FP16 and uniform low-bit quantization in the same pipeline.
Table~\ref{tab:modern_dense_full} gives the complete zero-shot results, extending the averages reported in the main text.
QuantaSpike remains close to uniform quantization on both model families under a four-step firing window.
The selected group size is particularly important for Llama-3-8B, where the final QuantaSpike configuration achieves $6.98$ WikiText-2 perplexity and a $71.25$ zero-shot average.
On Qwen3-8B, setting the admission gate to $\gamma=0.6$ yields a $69.41$ zero-shot average while retaining $n=3$ residual steps.

\begin{table*}[t]
\centering
\caption{
\textbf{Full results on newer dense LLMs.}
We report WikiText-2 perplexity and zero-shot accuracy.
Normalized accuracy is used for PIQA, ARC-Easy, ARC-Challenge, and HellaSwag; accuracy is used for WinoGrande.
All three methods are evaluated in the same pipeline.
}
\label{tab:modern_dense_full}
\scriptsize
\setlength{\tabcolsep}{3.5pt}
\renewcommand{\arraystretch}{1.05}
\begin{tabular*}{\textwidth}{@{\extracolsep{\fill}}llccccccc@{}}
\toprule
Model & Method & PPL $\downarrow$ & PIQA & ARC-Easy & ARC-Challenge & HellaSwag & WinoGrande & Avg. $\uparrow$ \\
\midrule
\multirow{3}{*}{Llama-3-8B}
& FP16 & 6.14 & 80.85 & 77.69 & 53.33 & 79.15 & 72.61 & 72.73 \\
& Uniform & 6.88 & 79.49 & 77.15 & 50.43 & 77.79 & 71.82 & 71.34 \\
& QuantaSpike & 6.98 & 80.15 & 77.05 & 50.11 & 77.47 & 71.47 & 71.25 \\
\midrule
\multirow{3}{*}{Qwen3-8B}
& FP16 & 9.71 & 77.75 & 80.93 & 56.57 & 74.93 & 67.72 & 71.58 \\
& Uniform & 10.08 & 75.90 & 76.14 & 53.41 & 73.72 & 65.90 & 69.01 \\
& QuantaSpike & 10.04 & 76.42 & 76.55 & 54.72 & 73.19 & 66.18 & 69.41 \\
\bottomrule
\end{tabular*}
\end{table*}

\subsection{Sensitivity to Group Size, Admission Gate, and Firing Window}

Table~\ref{tab:modern_sensitivity} reports the compact search used for the newer dense models.
Reducing $G$ from $64$ to $32$ consistently improves perplexity, indicating that finer gain alignment is more important than increasing the residual firing budget for these models.
Increasing $n$ from $3$ to $4$ at fixed $G=32$ and $\gamma=0.5$ gives only a small additional reduction, from $6.979$ to $6.958$ on Llama-3-8B and from $10.048$ to $10.046$ on Qwen3-8B.
We therefore retain $n=3$ in the main configuration to preserve the four-step firing window.
For Qwen3-8B, $\gamma=0.6$ provides the best balance among the tested admission thresholds; increasing it to $0.7$ slightly worsens perplexity.

\section{Energy Accounting Details}
\label{app:energy}

\subsection{Firing Budget}
\label{app:firing_budget}

We quantify the firing budget by counting nonzero LTIF spikes rather than the maximum number of available firing steps.
Let $p_o$ denote the fraction of activation values that emit an adaptive onset spike, and let $p_z$ denote the fraction of silent steps among all residual LTIF steps.
For a window containing $n$ residual steps and one optional onset step, the average number of nonzero events per activation is
  \begin{equation}
      \bar{m}=n(1-p_z)+p_o.
      \label{eq:appendix_event_budget}
  \end{equation}
The first term counts nonzero residual events, while the second accounts for onset events emitted by admitted outliers.
We use $n=3$ for the results reported here.
Each nonzero event triggers an ACC update for each associated synaptic connection, whereas a silent step triggers no accumulation.

To isolate the accumulation cost associated with this event count, we report the normalized activation-side arithmetic ratio
  \begin{equation}
      R_{\mathrm{act}}
      =
      \frac{\bar{m}E_{\mathrm{ACC}}}{E_{\mathrm{MAC}}},
      \label{eq:appendix_energy_ratio}
  \end{equation}
  where $E_{\mathrm{ACC}}=0.9\,\mathrm{pJ}$ and
  $E_{\mathrm{MAC}}=4.6\,\mathrm{pJ}$.
The reference is one MAC per activation--weight interaction.
This simplified ratio accounts for accumulation only; it excludes shift operations, memory access, and other execution overheads.
It is therefore an event-based arithmetic proxy, rather than a measurement of end-to-end hardware energy.
The bit-dependent ACC costs and shift costs used for the linear-energy projection are treated separately in the following subsection.

To quantify the contribution of zero-spike silence, we compare the actual nonzero-event count with an accounting reference that treats every residual step as active.
This reference retains the same onset rate, LTIF representation, and reconstructed values, but sets $p_z=0$ for accounting purposes, giving
  \begin{equation}
      \bar{m}_{\mathrm{all}}=n+p_o.
  \end{equation}
It does not modify the quantizer or constitute a separate inference experiment; perplexity is unchanged by construction.
The reduction in counted events is
  \begin{equation}
      \bar{m}_{\mathrm{all}}-\bar{m}=np_z.
  \end{equation}

Table~\ref{tab:appendix_silence} reports the firing statistics used in the energy analysis.
On OPT-1.3B, $56.37\%$ of residual steps are silent, reducing the average counted events per activation from $3.03$ to $1.34$.
On Llama-2-7B, $30.92\%$ of residual steps are silent, reducing the count from $3.01$ to $2.08$.
The corresponding normalized arithmetic ratios decrease from $0.593$ to $0.262$ and from $0.588$ to $0.407$, respectively.
These results show how data-dependent silence reduces the actual event count within the four-step firing window.

  \begin{table}[t]
      \centering
      \caption{
      \textbf{Effect of zero-spike silence on event accounting.}
      All residual steps counted is an accounting reference that
      treats silent residual steps as active; it preserves the
      same representation and perplexity as QuantaSpike.
      $p_o$ is the onset-event rate, $p_z$ is the silent residual-step
      rate, $\bar{m}$ is the average counted events per activation,
      and $R_{\mathrm{act}}$ is the normalized accumulation-cost ratio.
      }
      \label{tab:appendix_silence}
      \scriptsize
      \setlength{\tabcolsep}{4pt}
      \renewcommand{\arraystretch}{1.08}
      \begin{tabular}{llcccc}
          \toprule
          Model & Accounting
          & $p_o$ & $p_z$ & $\bar{m}$ & $R_{\mathrm{act}}$ \\
          \midrule
          \multirow{2}{*}{OPT-1.3B}
          & All residual steps counted
          & 2.90\% & 0.00\% & 3.03 & 0.593 \\
          & QuantaSpike (nonzero only)
          & 2.90\% & 56.37\%
          & \textbf{1.34} & \textbf{0.262} \\
          \midrule
          \multirow{2}{*}{Llama-2-7B}
          & All residual steps counted
          & 0.77\% & 0.00\% & 3.01 & 0.588 \\
          & QuantaSpike (nonzero only)
          & 0.77\% & 30.92\%
          & \textbf{2.08} & \textbf{0.407} \\
          \bottomrule
      \end{tabular}
  \end{table}

  \subsection{Linear-Energy Projection}

The absolute linear-energy comparison in Fig.~\ref{fig:linear_energy} uses the operation-level accounting adopted by the prior linear-layer analyses cited in Section~\ref{sec:energy_analysis}.
This projection concerns one linear transformation and should be interpreted as an analytical estimate rather than a measurement of end-to-end hardware power.

For QuantaSpike, each nonzero residual LTIF event consists of one power-of-two shift and one 4-bit ACC update.
An admitted outlier additionally emits one onset event, whose response is accumulated with the corresponding 5-bit ACC cost. We normalize all operation costs by the energy of a 4-bit ACC and write the normalized QuantaSpike event cost as
  \begin{equation}
      C_{\mathrm{QS}}(M)
      =
      n(1-p_z)(1+\alpha_{\mathrm{shift}})
      +
      p_o(\rho_5+\alpha_{\mathrm{shift}}),
      \label{eq:appendix_shift_cost}
  \end{equation}
where $n$ is the number of residual firing steps, $p_z$ is the fraction of silent residual steps, and $p_o$ is the fraction of values that emit an adaptive onset spike. The quantity
  \[
  \rho_5
  =
  \frac{E_{\mathrm{ACC}}^{5}}{E_{\mathrm{ACC}}^{4}}
  \]
  denotes the relative cost of a 5-bit ACC, and
  \[
  \alpha_{\mathrm{shift}}
  =
  \frac{E_{\mathrm{shift}}}{E_{\mathrm{ACC}}^{4}}
  =
  0.2
  \]
denotes the normalized cost of a power-of-two shift. The residual term therefore accounts for nonzero 4-bit LTIF spikes, whereas the onset term accounts for the additional 5-bit event emitted by admitted outliers.

For the SpikeQuant reference, we use the same operation-energy accounting as the source linear-layer analysis.
Let $\rho_M$ denote the reported salient-value ratio for model $M$, and let $\mu_{\mathrm{n}}$ and $\mu_{\mathrm{o}}$ denote the normalized MAC costs for the normal and salient-value paths, respectively.
The corresponding SpikeQuant cost is
  \begin{equation}
      C_{\mathrm{SQ}}(M)
      =
      (1-\rho_M)(1+\mu_{\mathrm{n}})
      +
      \rho_M(\rho_5+\mu_{\mathrm{o}}),
      \label{eq:appendix_spikequant_cost}
  \end{equation}
where the numerical values used in the projection are $\mu_{\mathrm{n}}=8.66$ and $\mu_{\mathrm{o}}=9.24$.
These values follow the normalized operation costs in the reference analysis, while $\rho_M$ follows the corresponding reported salient-value ratio.

We project the QuantaSpike energy by scaling the reported SpikeQuant linear-transformation energy according to the ratio of the two normalized operation costs:
  \begin{equation}
      E_{\mathrm{QS}}^{\mathrm{linear}}(M)
      =
      E_{\mathrm{SQ}}^{\mathrm{linear}}(M)
      \frac{C_{\mathrm{QS}}(M)}
           {C_{\mathrm{SQ}}(M)}.
      \label{eq:appendix_linear_projection}
  \end{equation}
Here $E_{\mathrm{SQ}}^{\mathrm{linear}}(M)$ is the reported SpikeQuant energy for one linear transformation, and $E_{\mathrm{QS}}^{\mathrm{linear}}(M)$ is the corresponding QuantaSpike projection under the same accounting model.

For OPT-1.3B and Llama-2-7B, the values of $p_o$ and $p_z$ are obtained from the measured firing statistics in Table~\ref{tab:appendix_silence}.
For OPT-2.7B and Llama-2-13B, we reuse the measured statistics from the smaller model in the same model family because exact firing diagnostics were not recorded for those models.
The resulting QuantaSpike values are therefore marked as estimates in Table~\ref{tab:appendix_linear_energy}.

Table~\ref{tab:appendix_linear_energy} lists the numerical values used to generate Fig.~\ref{fig:linear_energy}. 
The projected QuantaSpike linear energy is approximately $80.0\%$ lower than SpikeQuant on OPT models and $67.1\%$ lower on Llama-2 models under this operation-level accounting. These percentages describe the estimated energy of a single linear transformation. 
They do not represent measured end-to-end energy, and they do not include attention matrix multiplications, memory-system overheads, or hardware-specific scheduling effects.

\begin{table*}[t]
\centering
\caption{
\textbf{Linear-energy values used in Fig.~\ref{fig:linear_energy}} ($\mu$J).
Prior-method values follow the corresponding reported linear-layer analysis.
QuantaSpike values are obtained from the operation-level projection above.
$^{*}$ indicates an estimate using same-family firing statistics. Lower is better.
}
\label{tab:appendix_linear_energy}
\scriptsize
\setlength{\tabcolsep}{4.0pt}
\renewcommand{\arraystretch}{1.05}
\begin{tabular*}{\textwidth}{@{\extracolsep{\fill}}lcccc@{}}
\toprule
Method & OPT-1.3B & OPT-2.7B & Llama-2-7B & Llama-2-13B \\
\midrule
Quantization & 4.732 & 7.393 & -- & -- \\
OneBit & 13.237 & 20.683 & -- & -- \\
LRQuant & 13.690 & 21.390 & -- & -- \\
GPTQ & 14.227 & 22.229 & -- & -- \\
SmoothQuant & -- & -- & 18.927 & 29.573 \\
OmniQuant & -- & -- & 18.927 & 29.573 \\
Atom & -- & -- & 21.454 & 33.398 \\
SpikeLLM & -- & -- & 20.668 & 34.887 \\
Uniform (ours) & 4.753 & 7.426 & 18.948 & 29.606 \\
SpikeQuant & 2.790 & 4.358 & 11.082 & 17.318 \\
Kirin & 2.851 & 4.424 & 11.363 & 17.688 \\
\textbf{QuantaSpike} & \textbf{0.558} & \textbf{0.872}$^{*}$ & \textbf{3.643} & \textbf{5.695}$^{*}$ \\
\bottomrule
\end{tabular*}
\end{table*}

\section{Additional Ablation Results}
\label{app:ablation}

\subsection{Adaptive Onset Spike}

The onset-spike ablation tests whether outliers can be represented by the residual LTIF dynamics alone.
Removing the onset response yields only a modest reduction in the event ratio, from $0.262$ to $0.230$ on OPT-1.3B and from $0.407$ to $0.382$ on Llama-2-7B.
The reconstruction penalty is substantially larger: outlier MSE increases by $16.3\times$ on OPT-1.3B and $6.3\times$ on Llama-2-7B.
The corresponding perplexity degradation shows that the onset event is not redundant firing overhead.
It provides the additional membrane correction required to return admitted outliers to the operating range of the shared residual LTIF dynamics.

\begin{table*}[h]
\centering
\caption{
\textbf{Ablation of the adaptive onset spike.}
Removing the onset response saves a small event budget but sharply increases outlier reconstruction error.
}
\label{tab:appendix_onset}
\scriptsize
\setlength{\tabcolsep}{2.7pt}
\renewcommand{\arraystretch}{1.04}
\begin{tabular*}{\textwidth}{@{\extracolsep{\fill}}llccccc@{}}
\toprule
Model & Variant & PPL & Outlier MSE & $p_o$ & $\bar{m}$ & Ratio \\
\midrule
\multirow{2}{*}{OPT-1.3B}
& QuantaSpike & \textbf{16.380} & \textbf{0.00326} & 2.90\% & 1.337 & 0.262 \\
& w/o onset & 18.784 & 0.05322 & 0.00\% & \textbf{1.178} & \textbf{0.230} \\
\midrule
\multirow{2}{*}{Llama-2-7B}
& QuantaSpike & \textbf{5.779} & \textbf{0.00334} & 0.77\% & 2.082 & 0.407 \\
& w/o onset & 6.039 & 0.02093 & 0.00\% & \textbf{1.950} & \textbf{0.382} \\
\bottomrule
\end{tabular*}
\end{table*}

\subsection{Magnitude Gate}

\begin{wraptable}{r}{0.48\linewidth}
\centering
\caption{
\textbf{Ablation of the magnitude gate on Llama-2-7B.}
MAD-only admission spends more onset events but severely degrades reconstruction.
}
\label{tab:appendix_gate}
\scriptsize
\setlength{\tabcolsep}{4.1pt}
\renewcommand{\arraystretch}{1.04}
\begin{tabular}{lcccc}
\toprule
Variant & PPL & $p_o$ & $\bar{m}$ & Ratio \\
\midrule
QuantaSpike & \textbf{5.779} & \textbf{0.77\%} & \textbf{2.082} & \textbf{0.407} \\
w/o magnitude gate & 39.470 & 4.40\% & 2.270 & 0.445 \\
\bottomrule
\end{tabular}
\end{wraptable}

MAD alone identifies values that are unusual relative to their channel distribution, but not every such value lies outside the gain-aligned LTIF range.
As shown in Table~\ref{tab:appendix_gate}, removing the magnitude gate increases the onset rate from $0.77\%$ to $4.40\%$ on Llama-2-7B, yet perplexity rises sharply from $5.78$ to $39.47$.
This behavior is therefore not an energy--accuracy trade-off.
Without the magnitude gate, false admissions alter the initial membrane correction for values that are already representable by the residual LTIF prototypes, thereby disrupting the calibrated reconstruction path.
The magnitude gate thus serves as an admission mechanism rather than a conventional sparsity heuristic.

\section{Diagnostic Extension to Mixture-of-Experts Models}
\label{app:moe}

We evaluate whether the LTIF representation can be applied to routed expert activations using Qwen3-30B-A3B.
Unlike a dense model, an MoE model exposes each expert to only the tokens selected by the router.
Expert-wise calibration is therefore uneven: frequently selected experts receive many calibration values, whereas rarely selected experts may receive too few valid samples to estimate stable gains and reset prototypes.

We consider two scopes.
The attention setting applies QuantaSpike to attention and shared dense projections while keeping routed experts and the router in high precision.
The expert setting additionally applies QuantaSpike to routed expert projections, while the router remains in high precision.
To test whether calibration coverage is the limiting factor, we also use a more diverse calibration set that activates a larger number of expert instances.

Table~\ref{tab:appendix_moe} reports a small-scale WikiText-2 diagnostic with eight sequences of length $512$.
Quantizing attention and shared dense projections preserves the FP16 perplexity on this split.
Extending QuantaSpike to routed experts increases perplexity from $11.97$ to $12.25$.
The diverse calibration set substantially reduces the number of skipped expert calibration entries but does not change perplexity at this evaluation scale.
These results establish functional compatibility with routed expert computation, but they should not be interpreted as a full WikiText-2 benchmark.

\begin{table*}[h]
\centering
\caption{
\textbf{Diagnostic MoE results on Qwen3-30B-A3B.}
The router remains in high precision.
Evaluation uses eight WikiText-2 sequences of length $512$; lower perplexity is better.
}
\label{tab:appendix_moe}
\scriptsize
\setlength{\tabcolsep}{2.8pt}
\renewcommand{\arraystretch}{1.04}
\begin{tabular*}{\textwidth}{@{\extracolsep{\fill}}llccc@{}}
\toprule
Variant & Quantized scope & Calibrated & Skipped & PPL \\
\midrule
FP16 & None & -- & -- & 11.970 \\
Attention & Attn. + shared dense & 192 & 0 & \textbf{11.900} \\
Experts & Routed experts & 10,578 & 7,854 & 12.254 \\
Experts + diverse calib. & Routed experts & 13,488 & 4,944 & 12.254 \\
\bottomrule
\end{tabular*}
\end{table*}

The skipped entries arise from incomplete routing coverage during calibration rather than from the LTIF quantizer itself.
A larger calibration corpus or routing-aware sample selection would improve expert coverage, but a full MoE study is outside the scope of the dense-model evaluation in the main paper.

\section{Limitations and Future Work}
\label{app:limitations_future}

\paragraph{Limitations.}
Our main evaluation focuses on dense decoder-only LLMs, including OPT, Llama-2,
Llama-3, and Qwen3. The MoE experiment in Appendix~\ref{app:moe} is a small
diagnostic rather than a full routed-expert benchmark: expert activation counts
are uneven, several expert-specific calibration entries remain incomplete, and
the evaluation uses only a short WikiText-2 split. We do not evaluate
vision-language models, where visual-token statistics, modality-specific
outliers, and cross-modal projection layers may require different admission and
calibration rules.

\paragraph{Future work.}
Future work will develop routing-aware calibration for MoE models, with sample
allocation that provides sufficient coverage for rarely selected experts and
separate statistics for shared and expert projections. We also plan to extend
LTIF to vision-language models by studying modality-aware gains and onset
admission for visual and textual activations.

\end{document}